\documentclass[11pt]{article}
\usepackage{coling2020}
\usepackage{times}
\usepackage{url}
\usepackage{latexsym}
\usepackage{multirow}
\usepackage{booktabs,subcaption,amsfonts,dcolumn} 
\usepackage[T1]{fontenc}

\usepackage[textsize=footnotesize]{todonotes}

\definecolor{Gray}{gray}{0.95}

\newcommand{\corpus}{\texttt{ApposCorpus}} 

\colingfinalcopy 

\title{The ApposCorpus:\\ A new multilingual, multi-domain dataset for factual appositive generation}

\author{Yova Kementchedjhieva \\
  University of Copenhagen\\
  {\tt yova@di.ku.dk} \\\And
  Di Lu \\
  Dataminr, Inc.\\
  {\tt dlu@dataminr.com} \\\And
  Joel Tetreault \\
  Dataminr, Inc.\\
  {\tt jtetreault@dataminr.com}}

\date{}

\begin{document}
\maketitle
\begin{abstract}
News articles, image captions, product reviews and many other texts mention people and organizations whose name recognition could vary for different audiences. In such cases, background information about the named entities could be provided in the form of an appositive noun phrase, either written by a human or generated automatically.
We expand on the previous work in appositive generation with a new, more realistic, end-to-end definition of the task, instantiated by a dataset that spans four languages (English, Spanish, German and Polish), two entity types (person and organization) and two domains (Wikipedia and News). 
We carry out an extensive analysis of the data and the task, pointing to the various modeling challenges it poses. The results we obtain with standard language generation methods show that the task is indeed non-trivial, and leaves plenty of room for improvement.  
\end{abstract}

\section{Introduction}

\blfootnote{

    %
    %
     \hspace{-0.65cm}  
     This work is licensed under a Creative Commons 
     Attribution 4.0 International Licence.
     Licence details:
     \url{http://creativecommons.org/licenses/by/4.0/}.
    %
    %
}

News articles, image captions, product reviews and many other texts mention people and organizations, whose name recognition could vary for different audiences. A piece of news, for example, may concern people and organizations that are known locally, but are not necessarily well-recognized on a global level. In such cases, news pieces targeted at a wider audience would provide background information about the entity in focus, often in the form of an \textit{appositive}. For example:
\begin{quote}
In March 2017 , Natalie Jaresko, \textit{former Minister of Finance in Ukraine}, was appointed as the board's executive director.
\end{quote}
It is unlikely that many people outside of Ukraine know the name Natalie Jaresko, so a foreign reader would likely benefit from the extra bit of information about her former occupation as justification for her new appointment. An appositive could also be less contextualized and provide information of more general importance, for example:

\begin{quote}
The conservation unit is in the Calhau bairro of São Luís, \textit{the state capital}.
\end{quote}

In general terms, appositives are phrases that appear next to a noun phrase and serve an explicative function \cite{bauer2017nominal}.
Adding such explanations to text is a multi-step process. First, one has to decide whether an entity mention needs an appositive. That may not be the case for entities that are sufficiently well-known
or that have been introduced earlier in the text. In case an appositive is indeed needed, the next step is to choose what information about the entity to disclose. If the information is to be of a factual nature, the writer needs to have prior knowledge of the entity, or access to an external knowledge resource--\newcite{kang2019pomo} found appositives to be frequently based on facts of particular relevance to the context of the mention. Lastly, the surface form of the appositive, well-fitted to the surrounding context, needs to be produced. Viewed from the perspective of NLP, appositive generation is therefore an interesting and challenging natural language generation problem that involves reasoning over facts from an external knowledge source, with reference to a given context.

The task of appositive generation, first introduced by \newcite{kang2019pomo}, is still in its early stages and data resources are limited. 
We expand on previous work in appositive generation with a new, more realistic, end-to-end definition of the task, instantiated by a dataset, \corpus,\footnote{Available at \url{https://yovakem.github.io/#ApposCorpus}.} that spans four languages (English, Spanish, German and Polish), two entity types (person and organization) and two domains (Wikipedia and News). While Wikipedia as a domain is curated for a world-wide audience and as such may not benefit much from appositive generation, we posit that it is a valuable source of abundant cross-lingual data which could be used as the basis for transfer learning. In addition to a large training set automatically sourced from Wikipedia, we therefore also introduce a gold standard test sourced from newswire, one of the true target domains for appositive generation \cite{kang2019pomo}. 


The next section of the paper outlines the theoretical framework behind the task of appositive generation. In Section~\ref{sec:silver_data} we describe the automatic data collection procedure used to obtain training data, and in Section~\ref{sec:gold_data} we detail the further manual validation performed to obtain quality data for cross-domain evaluation. The properties of the dataset are discussed in detail in Section~\ref{sec:data_analysis}. Sections~\ref{sec:experiments},~\ref{sec:results} and \ref{sec:analysis} describe the experiments we performed and the main findings from them. Section~\ref{sec:conclusion} concludes the paper and outlines avenues for future research.

\section{The task: Appositive generation}
\label{sec:appo}
\newcite{kang2019pomo} laid the groundwork for appositive generation and our work can be seen as an expansion of their efforts. Yet, we both rename the task and redefine it in more general terms. 

\subsection{Prior work}
\newcite{kang2019pomo} introduced the task of appositive generation. To date this is the only work on this task. They designed a data collection procedure where appositives are identified by locating instances of the \texttt{appos} dependency label \cite{nivre-etal-2020-universal} in parsed text, and used it to build a dataset of appositives for \textsc{person} entities in English news articles. The candidate appositives were cross-referenced with the WikiData knowledge base \cite{10.1145/2629489} through word matching, and only those appositives were included in the final dataset which matched a fact from WikiData.

More generally, appositive generation relates to work on joint fact selection and generation \cite{liang-etal-2009-learning,kim-mooney-2010-generative,angeli-etal-2010-simple,Konstas2013AGM}. 

\subsection{A shift in terminology}
\newcite{kang2019pomo} actually called the phrases in question \textit{post-modifiers}, rather than appositives. 
The linguistic term \textit{post-modifier} can be seen as subsuming appositives, but it is much broader, including also prepositional, non-finite and dependent clauses that appear in postposition. Meanwhile, appositives come in two forms, nominal appositives, where a single noun identifies or qualifies another noun, e.g. President \textit{Washington}, and explicative appositives, where a pronoun, an infinitive or a noun phrase is used to explain or specify the status of a noun \cite{bauer2017nominal}. Explicative appositives are further characterized as \textit{non-essential}, meaning that they are not integral to the grammatical or semantic well-formedness of the sentence it appears in, and as such are often delimited from the rest of the sentence by punctuation marks \cite{traffis_2019}. For the purposes of providing background information about named entities, we are in particular interested in \textit{explicative appositive noun phrases}, and that is what we refer to as an \textit{appositive} throughout this work. 

\subsection{Expanding the task definition}
\label{ssec:issues}
\begin{figure}
    \centering
    \includegraphics[width=\linewidth]{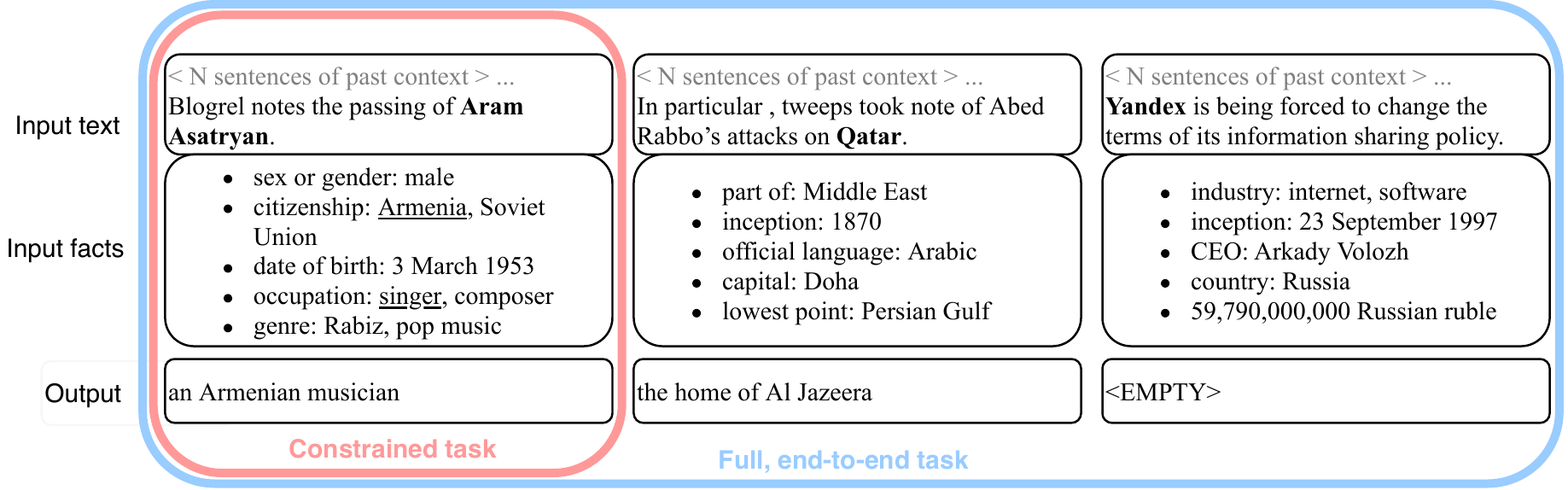}
    \caption{Illustration of the task in a \textit{constrained} setting, where an appositive is always due and the facts in it are always available in the knowledge base; and in a \textit{full, end-to-end} setting, where a decision has to be made as to whether or not to generate an appositive, and the facts in the appositive may be missing from the knowledge base. The entity in focus is shown in bold, the relevant facts are underlined (where available), and the $<$EMPTY$>$ tag means no appositive is needed. Optionally, previous context can be included in the input, e.g. the three previous sentences--this is not shown in the figure.}
    \label{fig:task}
\end{figure}

Being built with reference to WikiData, the dataset of \newcite{kang2019pomo} creates the illusion that all facts necessary to generate an appositive are available in the knowledge base. \newcite{balaraman2018recoin} studied the relative completeness of WikiData entries and found gaps to be the norm rather than an exception. Moreover, the dataset of \newcite{kang2019pomo} only includes positive samples, i.e. instances where an appositive is due. A more realistic scenario would also require the model to choose whether or not to add an appositive to a given entity mention. \corpus~ is \textit{not} constrained by WikiData in terms of fact matching, and contains positive \textit{and} negative samples, i.e. instances of empty appositives. See Figure~\ref{fig:task} for an illustration. Moreover, it is multilingual and covers both \textsc{person} and \textsc{organization} named entities. We built this dataset primarily based on text from Wikipedia, chosen for its rich cross-lingual coverage.

\section{Dataset collection: Wikipedia}
\label{sec:silver_data}
We used the March 2020 Wikipedia dump\footnote{\url{https://dumps.wikimedia.org/}} for English, Spanish, German and Polish, which we parsed with WikiExtractor,\footnote{\url{https://github.com/attardi/wikiextractor}} preserving internal links.\footnote{These links point to other pages on Wikipedia and allow us to identify the Wikidata entry for the given named entity.} The choice of these particular languages was mostly based on the availability of good dependency parsers. Since dependency parsing is an integral step of the data collection process \cite{kang2019pomo}, it has to be as precise as possible, to maximize the quality of the outcome.\footnote{We only considered parsers with labeled accuracy score over $90.0$}  Below, we describe in detail the data collection procedure. 

\subsection{Preprocessing}
We processed every article as follows: 
(1) tokenize the text and segment sentences; 
(2) normalize mentions of the entity in the article's title and annotate them with internal links; (3) identify sentences which contain a linked named entity listed as an instance of type \textit{human} or of (a subclass of) type \textit{organization} in WikiData (corresponding to the \textsc{person} and \textsc{organization} named entity types); 
(4) run a dependency parser 
on these sentences. 
Steps (1) and (4) were performed with Stanza \cite{qi2020stanza}.

\subsection{Detecting  appositives}
Any instance of the \texttt{appos} label that depends on a linked named entity and is separated from it with a comma or an opening parenthesis was considered a valid candidate. In this case, we recorded the source sentence, replacing the appositive with special token \texttt{<appos>}, as input data, and the appositive as a target. The beginning of the appositive was taken to be the first token after the comma or opening parenthesis, and the end is taken to be the last token before the next comma/semicolon/full stop (if beginning was marked by a comma) or closing parenthesis (if beginning was marked by an opening parenthesis). We discarded any commas and parenthesis surrounding the appositive, but kept semicolons and full stops as part of the input sentence. We also recorded the three preceding sentences from the article and one following sentence.
Similarly to \newcite{kang2019pomo}, we process one appositive per sentence, i.e. if there are multiple appositives in a sentence, we select the first one and do not consider the rest.

Appositives containing just dates (usually the date of birth and/or death of a person) are ubiquitous across Wikipedia articles to the point that they constitute up to 30\% of the data samples that we get with the procedure described above. 
We reduced this imbalance in the data by downsampling this type of appositives to only 10\% of its occurrences. 

\subsection{Negative samples}

We added negative samples to the dataset, matching the number of positive ones. They were drawn according to the following criteria: (1) there is a \textsc{person} or \textsc{organization} entity in the sentence, (2) it is not followed by a comma or opening parenthesis, and (3) the rest of the sentence does not contain an appositive dependent on the \textsc{person} or \textsc{organization} entity. Condition (2) was used to reduce the chance of including instances of appositives that were not correctly tagged as such by the parser (recall that appositives are often delimited from the rest of the sentence by a punctuation), while (3) was used to ensure that we did not include instances that contain a non-essential appositive, which the author had failed to delimit by any punctuation. In negative samples the input is the original source sentence with an added token \texttt{<appos>} just after the \textsc{person}/\textsc{organization} entity, and the target is a special \texttt{<EMPTY>} token. 



The procedure described above was used to collect \textit{training} data for factual appositive generation. As it is our goal to study the potential of a cross-domain approach to appositive generation, the \corpus~also contains an out-of-domain test set, sourced from newswire.

\section{Dataset collection: News}
\label{sec:gold_data}

We sourced our data for cross-domain evaluation from the news domain, following previous work \cite{kang2019pomo}, using these news corpora: Global Voices (English, Spanish, German, Polish) \cite{TIEDEMANN12.463}, News Commentary (English, Spanish, German) \cite{TIEDEMANN12.463} and Paralela \cite{pkezik2016exploring}.  


\subsection{Entity linking} Unlike Wikipedia, where entities are explicitly linked to WikiData entries through internal links, here, we had to perform additional entity linking. We did so in the following manner: (1) extract candidates from WikiData based on exact match between the full span of the named entity and all aliases of entities in the respective subset of WikiData (instances of type \textit{human} if NER label is \textsc{person} , else \textit{organization}), (2) obtain relative alias frequency distributions from Wikipedia, and (3) the candidate entity with the highest relative frequency given the alias is selected. We chose to use this prior-based method instead of a modeling approach since off-the-shelf entity linkers were not available for all the languages involved. 
Candidate appositives were then identified as described in 3.2. 

As errors could occur both in the entity linking and in the appositive detection, we hired manual annotators to verify the output of the two procedures (see details in Appendix~\ref{manual}). The News portion of the \corpus~ is therefore gold standard and will serve for stable, accurate evaluation. 

\subsection{Negative samples} We added $1,000-n$ negative samples to each subset of the data, where $n$ is the number of positive samples.
The exact ratio of positive samples in each test set is reflected in the \texttt{always yes} baseline shown in the Results section (see Figure~\ref{fig:gold}a), a dummy baseline which always predicts an appositive. 

\section{Data analysis}
\label{sec:data_analysis}
This section outlines some findings on the properties of our dataset, based on general statistics and WikiData cross-referencing. The procedure described above yielded less than four thousand samples of \textsc{org} appositives for Polish, so this subset is omitted from the \corpus. 

\begin{table}
    \small
    \begin{subtable}[l]{0.5\textwidth}
    \centering
    \begin{tabular}{clllll}
          && en & es & de & pl \\
          \hline
         \multirow{4}{*}{\rotatebox{90}{\textsc{per}}} & Size &  559k & 164k & 269k & 14k \\
         &Length&3.4&4.16&2.7&2.67\\
         & WD (\%) & 25.5 & 28.1& 21.2 & 22.3\\
         \hline
         \multirow{4}{*}{\rotatebox{90}{\textsc{org}}} & size & 612k & 104k & 333k & -\\
         &Length&2.19&4.09&1.64&-\\
         & WD (\%) & 27.8 & 24.9& 22.2 & -\\
    \end{tabular}
    \caption{Wikipedia data}
    \label{tab:silver_data_stats}
    \end{subtable}
    \begin{subtable}[l]{0.5\textwidth}
    \centering
    \begin{tabular}{clllll}
          && en & es & de & pl \\
          \hline
         \multirow{4}{*}{\rotatebox{90}{\textsc{per}}} & size &  1k & 1k & 1k & 1k\\
         & Length & 4.07&3.51&3.43&2.32\\
         & WD (\%) & 29.3& 30.8 & 21.7 & 20.7 \\
         \hline
         \multirow{4}{*}{\rotatebox{90}{\textsc{org}}} & Size &  1k & 1k & 1k & -\\
         & Length & 3.31 & 3.08 & 1.87 & - \\
         & WD (\%) & 35.4& 30.0& 22.8 & -  \\
    \end{tabular}
    \caption{News data}
    \label{tab:gold_data_stats}
    \end{subtable}
    \caption{Dataset statistics. \textit{Size}: full dataset size, \textit{Length}: average appositive length, \textit{WD}: ratio of appositives matching a fact from WikiData.}
    \label{tab:data_stats}
\end{table}

\subsection{General statistics}
Table~\ref{tab:data_stats} lists some statistics about the two parts of the dataset, one based on Wikipedia (\textit{Wikipedia data}) and the other on news (\textit{News data}). Row \textit{Size} refers to the full size of the data as split into language and entity type. We further split each Wikipedia subset for training (70\%), validation (15\%) and testing (15\%).  
Size varies greatly across the data, with the Polish \textsc{person} subset being merely 2.5\% the size of the English \textsc{person} subset. This relates both to a difference in the Wikipedia sizes for these languages  (6M English articles v. 1.4M Polish articles) and to a difference in the frequency of use of appositives across the languages.  

Row \textit{Length} in Table~\ref{tab:data_stats} lists the average number of tokens per appositive, which varies from two to four tokens, and is generally lower for \textsc{organization} appositives than for \textsc{person} ones. 

\subsection{Cross-referencing with WikiData}

As discussed before, fact matching is not part of our data collection procedure, but at training time it would be beneficial to have access to a knowledge base such as WikiData, and to draw from it, when possible. So we extract the WikiData entries for all named entities in our dataset and perform word matching between facts and appositives in the following way: (1) tokenize the fact and the appositive, (2) remove stopwords, (3) measure token overlap. If the overlap is non-zero, we consider there to be a match and annotate the fact as \texttt{used}.\footnote{We experimented with other thresholds (2 and 3-word overlap) and with fuzzy matching, but found this method to work best.}

Cross-referencing the data with WikiData is also useful as an insight into the makeup of the data, albeit an insight that is biased the scope and completeness of the knowledge base.

\paragraph{Coverage} Row \textit{WD} 
in Table~\ref{tab:silver_data_stats} shows the rather low percentage of appositives from the Wikipedia dataset that are matched to at least one fact from WikiData: from 21.2 for German \textsc{person} appositives to 28.1 for Spanish \textsc{person} appositives. The numbers for the News test set, shown in Table~\ref{tab:gold_data_stats}, are mostly similar to those for the WikiData. Another way to view these percentages is as an effective upper bound on the performance of a model trained with WikiData as the source of knowledge. Further work in identifying other sources of facts for appositive generation and new means of integrating them into a model could therefore prove very fruitful. 

We manually inspected a random sample of 100 appositives from the English section of the Wikipedia dataset that were not matched to any fact from WikiData. In the majority of cases, the appositives concerned the occupation of a person, their position within an organization, their country of origin, or other type of information that is typically found in WikiData, but was missing for the given named entry.



\paragraph{Composition} We studied the composition of the data, as observed with reference to WikiData. We performed our analysis on all languages and found that similar trends hold cross-lingually, so here we discuss the English portion of the data only, and in the Appendix~\ref{composition} we include the corresponding tables for Spanish, German and Polish. 


We looked at the types of facts that were matched to appositives from the News data. For all fact types that constitute 3\% or more of all facts matched, we also looked at their frequency in the Wikipedia data. Results are shown in Table~\ref{tab:wikidata}. We see that half of the top fact types in the News dataset are also well-attested in the Wikipedia data, i.e. their relative frequency is 3\% or more. We can expect that knowledge concerning appositives based on these fact types would trivially transfer from one domain to the other. The low frequency for the remaining fact types (cf. \textit{has quality} and \textit{capital of}), on the other hand, poses a challenge whose solution would require deeper natural language understanding and, possibly, explicit domain transfer techniques.   

\begin{table}[]
    \centering
    \small
    \begin{tabular}{clllclll}
                         &Fact type&News (\%)&Wiki(\%)&&Fact type&News (\%)&Wiki(\%)\\
                         \hline                         
                         \multirow{7}{*}{\rotatebox{90}{\textsc{per}}}&  position held & 20.9 & 9.4 & \multirow{7}{*}{\rotatebox{90}{\textsc{org}}} & instance of&23.1&10.9\\
                          & occupation&15.9&10.6&& official website & 6.3&6.2\\
                          & citizenship&10.1&4.3 && country&5.9&3.3\\
                          & member of party&7.6&1.9&& member of&4.2&2.4\\
                          & award received&5.2&3.9&& subsidiary&3.5&2.1\\
                          & nominated for&3.6&0.4&& capital of&3.2&0.1\\
                          & educated at&3.1&3.1&& has quality&3.0&0.0\\
                         \hline

    \end{tabular}
    \caption{Top fact types. English. }
    \label{tab:wikidata}
\end{table}

\section{Experiments}
\label{sec:experiments}


To show how the new task formulation can be used,  
we experiment with three established language generation methods: the main method of \newcite{kang2019pomo}, which we refer to as \texttt{base}; an extension of \texttt{base}  with external knowledge injected through embeddings with knowledge-base grounding (\texttt{KB}); and a model enhanced with an explicit copy mechanism (\texttt{copynet}, \newcite{gu-etal-2016-incorporating}). Notice that our goal here is not to build the best model for this task, but to develop reasonable models which can serve as baselines for future work in this area.\footnote{We also experimented with a transformer architecture, but encountered optimization problems. See details in Appendix~\ref{transformer}.}


\subsection{Architectures}
\paragraph{LSTM baseline, \texttt{base}}
\newcite{kang2019pomo} introduced an LSTM-based encoder-decoder architecture with an auxiliary objective used to guide the attention of the decoder towards the WikiData facts that were matched during the data collection process. Input sentences and facts are represented with the same word embeddings and encoded by separate biLSTMs. The decoder is initialized with the encoding of the input and attends over the encodings of the facts. Our only modification here is to add a ``None of the above'' item to the list of facts about an entity and point the attention to that when no other fact was matched or the appositive was empty (i.e. for negative data instances).

\paragraph{LSTM with external knowledge, \texttt{KB}}
External knowledge can be beneficial to a better understanding of the context and how it relates to the different facts known about an entity. Here, we use the same architecture as above, but initialize the embedding matrix of the model with the NTEE (Neural Text-Entity Encoder) word embeddings,
trained on Wikipedia with WikiData grounding \cite{yamada-etal-2017-learning}. They aim to represent a text and its relevant entities close to each other. We deem these embeddings suitable for our modeling setup, where input text and facts are represented in a shared space. Unfortunately, the NTEE embeddings are available only for English. So we used word-level translation to ``project'' them to Spanish, German and Polish. See more details in Appendix~\ref{projection}.    This approach is likely to introduce some noise, but we only use the projection to initialize the embedding matrix which is then further trained. So any signal coming from the embeddings can be used by the model and any noise can be filtered out during training. 

\paragraph{LSTM with a copy mechanism,  \texttt{copynet}}
Motivated by the observation that there is an overlap of at least one token between WikiData facts and appositives for about 25\% of the datapoints in our dataset, we experiment with a method that allows the decoder to copy tokens directly from the input: Copynet \cite{gu-etal-2016-incorporating}. \newcite{kang2019pomo} correctly point out that in their constrained data setting, where data points were selected based on word overlap with WikiData facts, using a copy mechanism would result in double-counting, i.e. artificially boosted results. In our data setting, however, this is not the case.
\\

All three approaches are  end-to-end in the sense that we do not split up the classification task of whether or not to predict an appositive from the task of generating an appositive where it is due. As negative samples in the dataset have the special \texttt{<EMPTY>} token as target, the models are performing the classification task \textit{implicitly} by choosing whether to predict the \texttt{<EMPTY>} token or not.

Preliminary experiments with the \texttt{base}  architecture showed that the choice between providing the model with three sentences of preceding context, one or zero had little impact on its performance, so all results reported below use one sentence of preceding context, following \newcite{kang2019pomo}. Further details on the implementations and the hyperparameters we used can be found in Appendix~\ref{implementation}. 

\subsection{Evaluation}
We follow \newcite{kang2019pomo} in the choice of performance metrics for predictions over the positive instances in the data: we measure F1 score of the predicted bag-of-words excluding stopwords; BLEU \cite{10.3115/1073083.1073135} over n-grams, where $n=1,2,3$;\footnote{\newcite{kang2019pomo} also included four-grams, but seeing that the average length of an appositive across the different subsets of the data is 3.08 tokens, we exclude four-grams from consideration.} and METEOR \cite{denkowski-lavie-2014-meteor}, which supports stemming and synonymy only in English, Spanish and German, so these features are not used for Polish. We use accuracy to measure the models' ability to determine when an appositive is due. 

\section{Results}
\label{sec:results}

\begin{table}
    \centering
    \small
    \begin{tabular}{cllllll}
         Train setting&Test setting&Dataset&Acc(\%)&F1 & BLEU & MET. \\ 
         \hline
         \multirow{2}{*}{constrained}&
         \multirow{2}{*}{constrained}& ApposCorpus (News) & - & 19.61	&7.93	&9.12 \\
         && PoMo & - &  11.21&	3.44&	5.03 \vspace{1mm}\\
         \multirow{4}{*}{end-to-end}&\multirow{2}{*}{constrained}& ApposCorpus (News) & 95.0 & 10.76 & 3.39 & 4.41 \\
         &&PoMo&91.7 & 4.52	&0.57	&2.03\vspace{1mm}\\ 
         &end-to-end&ApposCorpus (News) &72.33 & 5.97 & 1.03 & 2.96\\

    \end{tabular}
    \caption{Generation of English \textsc{person} appositives in a constrained v. end-to-end train and test setting.}
    \label{tab:RQ1}
\end{table}

We view the results of our experiments from two angles: one concerns the expansion of the task definition we achieve with \corpus, from a constrained scenario to an end-to-end one; the other concerns the increased coverage of the dataset, which allows us to compare and contrast appositive generation across different languages and named entity types. 

\subsection{Constrained v. end-to-end scenario}
To draw a direct comparison to the work of \newcite{kang2019pomo}, in this subsection we focus on English \textsc{person} appositives, as this is the subset that was covered by their dataset, dubbed \texttt{PoMo}. We begin by replicating exactly their train and test settings, both constrained, using the model architecture they proposed, \texttt{base}. In the first two rows of Table~\ref{tab:RQ1}, the performance of the model is reported on both the constrained subset of our News test data and on the \texttt{PoMo} test set, constrained by design. There is a considerable difference in performance as measured on the two test sets. Since they were both drawn from the same domain, this difference may largely be due to one test set being gold standard and the other silver standard, which highlights the importance of having gold standard evaluation data. 

Using these results as a starting point, we consider two important factors in the shift from a constrained to an end-to-end setting: one concerns learning and the other, evaluation. 

\paragraph{Learning complexity} can be expected to increase in the end-to-end training setting, since the model has to learn not just what appositive to predict, but also whether or not to predict an appositive. Due to gaps in WikiData, the model also has to learn how to best handle instances of appositives based on unobserved facts. We demonstrate how these factors affect performance by comparing the \texttt{base} model trained in a constrained setting to one trained in an end-to-end setting. We measure the models' performance in a constrained test setting, to make the comparison fair to the former. Shifting from a constrained train setting (rows 1 and 2 of Table~\ref{tab:RQ1}) to an end-to-end setting (rows 3 and 4), we observe a drop in performance of around 50\% on all generation metrics. The model trained end-to-end does very well on choosing whether or not to predict an appositive (accuracy is 95\% for \corpus~ and 91.7\% for \texttt{PoMo}) , so we have to conclude that the lower generation scores are not a matter of predicting empty appositives, but rather of predicting worse appositives due to the increased learning complexity.

\paragraph{Evaluation} is another aspect to consider when comparing the constrained and full data settings. The quality of evaluation is key to understanding how well a model would perform if deployed in the real world. Constrained evaluation, however, only tells us how a model would do in an idealistic scenario, where all the facts about all the entities were indeed covered by a knowledge base. As this is not the case with WikiData \cite{balaraman2018recoin}, and with any existing knowledge base for that matter, it is important to evaluate models in a manner that reflect gaps in external knowledge sources. We report the performance of a model trained and tested in an end-to-end setting in the last row of Table~\ref{tab:RQ1}. Compared to the model's performance as measured on the constrained test set (row 3), these numbers are substantially lower. Yet, they are the numbers that most truly represent the performance of the base model, at least in terms of automatic evaluation. We return to this matter when analysing the model's performance in Section~\ref{sec:analysis}.


\subsection{Languages and entity types}

\begin{figure*}
    \centering

    \begin{subfigure}{0.39\linewidth}
    \resizebox{\linewidth}{!}{
    \includegraphics{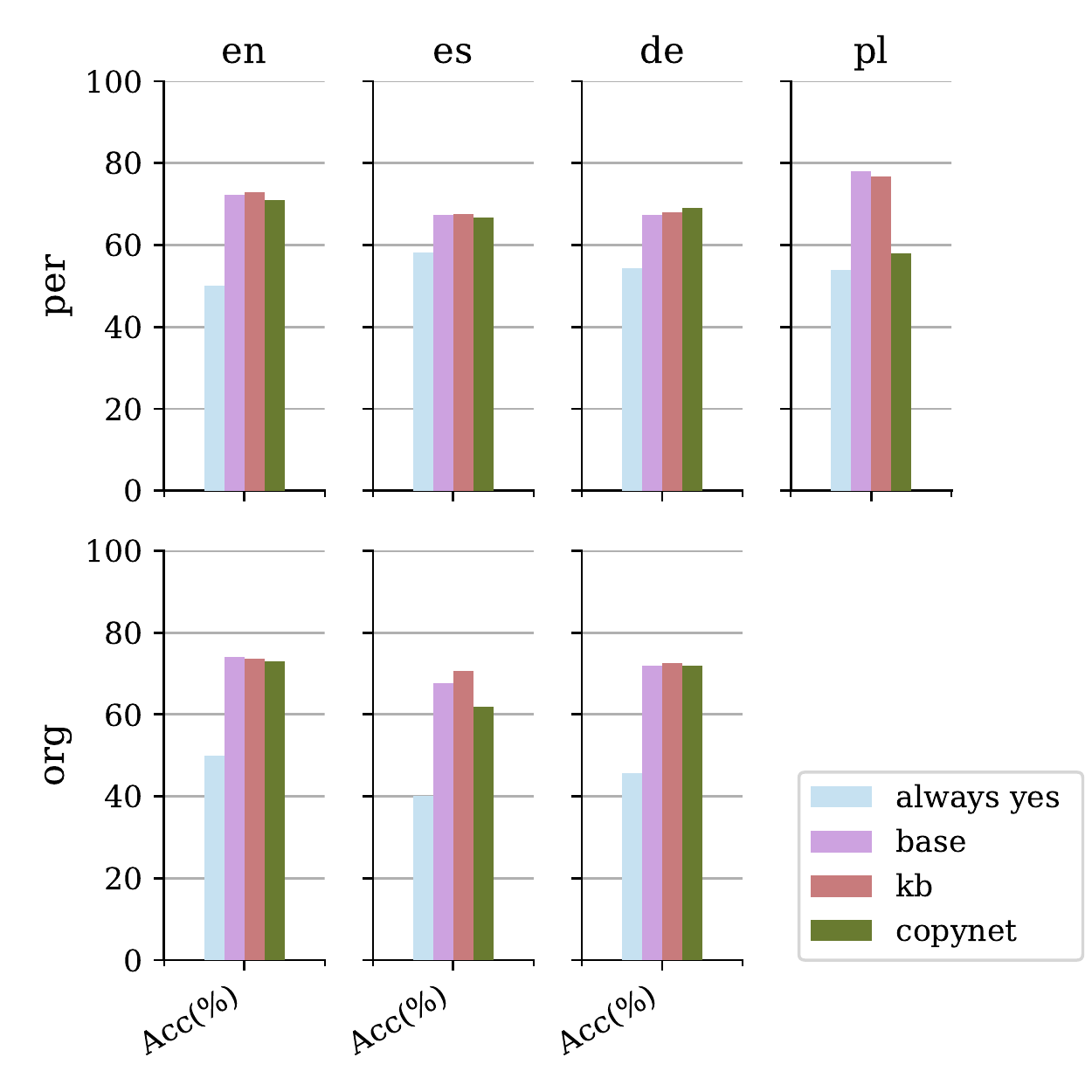}}
    \caption{}
    \end{subfigure}
    \begin{subfigure}{0.59\linewidth}
    \resizebox{\textwidth}{!}{
    \includegraphics{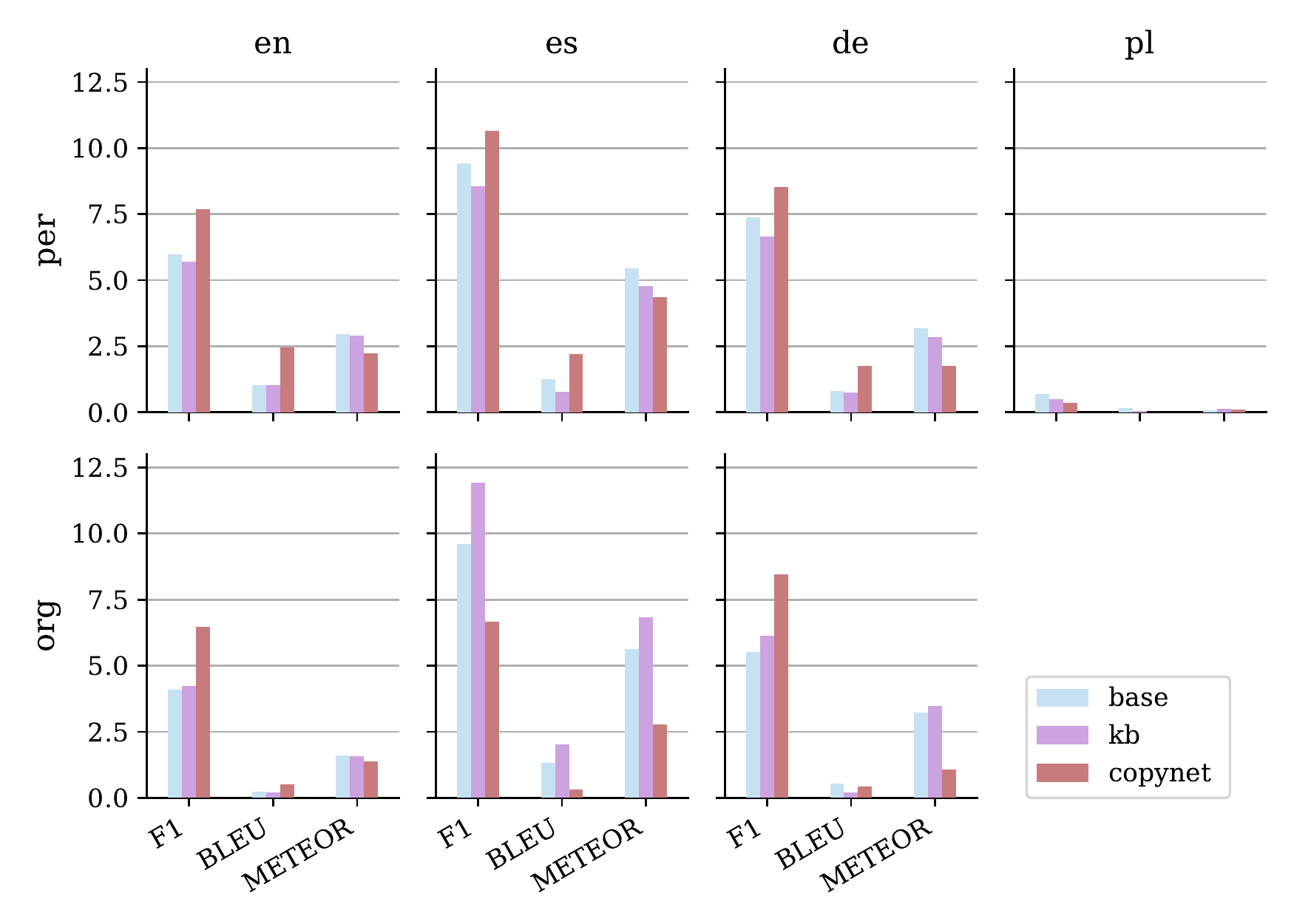}}
    \caption{}
    \end{subfigure}
    \caption{(a) Evaluation of (a) the models' ability to correctly decide when an appositive is due, (b) generated predictions for positive test instances. Measured on the News test set. }
    \label{fig:gold}
\end{figure*}

The full range of results on the News test set are shown in Figure~\ref{fig:gold}. Results are averaged over three models trained from different random initializations. We evaluate how well the models can detect when an appositive is needed (implicit classification) and how well it can perform the end-to-end task of classifying and generating a good appositive.

\paragraph{Implicit classification} of the positive and negative samples in the data appears to be roughly equally challenging across the two entity types and the four languages, as viewed across all three models (see Figure~\ref{fig:gold}a. One exception is Polish \textsc{person} appositives, where \texttt{base} and \texttt{kb} score higher than they do on the other subsets, while \texttt{copynet} barely beats the baseline. Since the models are not explicitly trained to perform this type of classification, it is encouraging to see that even in this setting, they can outperform the baseline (always predicting the positive class) by as much as 20\% in several cases.

\paragraph{Generation} is the more difficult aspect of the task, as shown in Figure~\ref{fig:gold}b. We see that, somewhat surprisingly considering the amounts of training data (see Table~\ref{tab:silver_data_stats}),  performance is not highest on English, but rather on Spanish. In line with the small amount of training data, on the other hand, performance on Polish is virtually non-existent. There are no clear differences between overall performance on \textsc{person} and \textsc{organization} appositives. Only in English, the latter seem to pose a greater challenge to all three models and according to all three metrics. Model comparison is not straightforward since the different metrics reveal different strengths and weaknesses in each approach. It does appear to be the case that injecting external knowledge through pretrained knowledge-base embeddings (\texttt{kb}) is beneficial to the prediction of \textsc{organization} appositives and somewhat harmful to the prediction of \textsc{person} appositives. Since the differences between the three methods are not consistent across all languages, named entity types and metrics, we cannot conclusively say which method is best, but we do note that \texttt{copynet} scores high on the most metrics, languages and entity types. To better understand the performance of this model, we stepped away from automatic generation metrics, which are known to suffer from certain biases and can be difficult to interpret, and we carried out an additional manual evaluation.

\section{Analysis}
\label{sec:analysis}
\subsection{Manual evaluation}
We used Amazon Mechanical Turk to carry out a ranking paradigm study on the predictions of \texttt{copynet} for English \textsc{person} and \textsc{organization} appositives. The annotators were shown a true appositive and a predicted appositive side-by-side (in the context of the input sentence) and were asked to express their preference towards either of these two, or their lack of preference as a third option. One example prompt is shown in Appendix~\ref{taste_test}. Five annotations were obtained per data instance, and we then took the majority vote as an indication of the overall preference. 
The results are shown in Table \ref{tab:manual_eval}.  
For  \textsc{person} appositives, the writer's choice (true) and the system's prediction 
were preferred at almost equal rates---this observation challenges the numbers obtained with the automatic evaluation metrics, as it suggests that the predicted appositives were not necessarily of poor quality. A bigger gap was observed between true and system-generated \textsc{organization} appositives, where the crowdworkers preferred the original appositive 66.0\% of the time. The lower preference for \textsc{organization} predictions is in line with the trend in the automatic results, where performance on English \textsc{organization} appositives was shown to be lower than that on English \textsc{person} appositives.  Notice, however, that even for \texttt{organization} appositives, annotators still showed preference for the predicted ones at a considerable rate. This suggests that the automatic metrics may have severely under-represented the abilities of the models. 


\begin{table}[]
    \centering
    \small
    \begin{tabular}{llllllll}
         & true & system & neutral && true & system & neutral \\
         \textsc{per}& 46.3\% & 47.3\% & 6.1\%  & \textsc{org}& 66.0\% & 26.5\% & 7.4\% \\
         
    \end{tabular}
   \caption{Results from the ranking paradigm study comparing true appositives to system-generated ones.}
    \label{tab:manual_eval}
\end{table}

\subsection{Qualitative analysis}
To better understand the source of error in the models' predictions, we manually inspected 50 data points from  the \textsc{person} subset and 50 data points from the \textsc{organization} subset, where a choice was made in favour of the true appositive over the predicted one. Half of the data points were instances where the true appositive was not empty, but the models predicted an empty appositive. The annotators seemed to strongly prefer non-empty appositives, possibly due to the fact that they where shown sentences without their original context, where the role of an entity might have been clarified at an earlier mention. Yet, that is not categorically so as seen in example 1 in Table~\ref{tab:examples}, where the predicted appositive is redundant in the given context, so the annotators preferred the true empty appositive. Other types of errors the models made were to predict appositives that concern the right piece of information but are too general (examples 2 and 3), to predict appositives based on the wrong piece of information (examples 4 and 5), and, specifically for \textsc{organization} appositives, to just repeat the named entity (example 6).   While the latter is the result of either suboptimal learning or noise in the data, the rest of the errors we saw point to the need for an approach with deeper understanding of the facts and their relevance to the context.

\newcolumntype{L}[1]{>{\raggedright\arraybackslash}p{#1}}
\newcolumntype{C}[1]{>{\centering\arraybackslash}p{#1}}
\newcolumntype{R}[1]{>{\raggedleft\arraybackslash}p{#1}}
\begin{table*}[]
    \small
    \resizebox{\textwidth}{!}{
    \begin{tabular}{L{0.005cm}L{11.7cm}L{2.9cm}}
        &Gold sentence & Prediction \\
        \hline
        1&In response to an April 9 court ruling declaring the military backed government of Frank Bainimarama \textcolor{blue}{\texttt{<EMPTY>}} to power illegally when he dissolved Parliament and deposed the government of Laisenia Qarase , the country ’s President nullified the Fiji 's constitution , fired the entire judiciary and appointment himself head of state and the armed forces. & the President of Fiji\\
        2&Artyom Loskutov, \textcolor{blue}{creator of the popular counter - culture art movement " Monstration "}, made waves on RuNet by signing a letter in support of Dmitry Kiselyov , a journalist who many consider to be Putin 's chief propagandist. & a Russian painter\\
        3&In a fatal blow to our already lackluster sources of entertainment , the Sudanese government has blocked access to YouTube,  \textcolor{blue}{the online video sharing Web site}. & platform \\
        3&Modi, \textcolor{blue}{a tech-savvy nationalist from the right-wing Bharata Janatiya Party}, has traveled the world to sell the idea of India as an emerging digital economy, making deals with the likes of Google and ( less successfully ) Facebook. & the Prime minister of India \\
        4&Some critics also highlighted the fact that Jabrailov is from Chechnya, \textcolor{blue}{a republic in the Northern Caucasus region of Russia where Muslim separatists fought two bloody wars against the Russian army}.&Chechnya\\
        5&He steers between Soukous , rhumba and RnB ” , and links to an interview with the singer on Radio Okapi,  \textcolor{blue}{the nationwide radio station sponsored by the UN and Fondation Hirondelle}. & the Democratic Republic of the Congo\\
        6&In particular , tweeps took note of Abed Rabbo 's attacks on Qatar, \textcolor{blue}{the home of Al Jazeera}. & Qatar\\
    \end{tabular}
    }
    \caption{Examples where true appositives were preferred over predicted ones by human annotators.}
    \label{tab:examples}
\end{table*}

\section{Conclusion}
\label{sec:conclusion}
\corpus~ targets factual appositive generation, a phenomenon frequently occurring in a range of textual domains.  It substantially extends the prior resources in the area by spanning four languages, two named entity types 
, and two domains.  This resource also allows the burgeoning field to investigate end-to-end appositive generation.  
Our manual and automatic evaluations with \corpus~ show that standard model architectures can approach the quality of human targets in specific cases but there is still room for improvement.
With this dataset, appositive generation can be studied in much more depth than previously possible, ultimately paving the way for novel NLP applications in the generation and writing space. The focus in future research, we believe, should fall on explicit methods for cross-domain learning, on richer knowledge sources, and on the development of test sets for new domains. 

\section{Acknowledgements}
We acknowledge the responsiveness and help of Jun Seok Kang and Robert L. Logan IV in discussions on the design choices and goals of their work on post-modifier generation and on how to accurately reproduce their experiments.  We also thank researchers from Dataminr: Saran Krishnasamy, Lin Nie, and Isabel Zhang for their help in setting up experiments and annotation.  Finally, we thank the three anonymous reviewers for their feedback and comments.

\bibliographystyle{coling}
\bibliography{coling2020}

\begin{thebibliography}{}

\bibitem[\protect\citename{Angeli \bgroup et al.\egroup
  }2010]{angeli-etal-2010-simple}
Gabor Angeli, Percy Liang, and Dan Klein.
\newblock 2010.
\newblock A simple domain-independent probabilistic approach to generation.
\newblock In {\em Proceedings of the 2010 Conference on Empirical Methods in
  Natural Language Processing}, pages 502--512, Cambridge, MA, October.
  Association for Computational Linguistics.

\bibitem[\protect\citename{Balaraman \bgroup et al.\egroup
  }2018]{balaraman2018recoin}
Vevake Balaraman, Simon Razniewski, and Werner Nutt.
\newblock 2018.
\newblock Recoin: relative completeness in wikidata.
\newblock In {\em Companion Proceedings of the The Web Conference 2018}, pages
  1787--1792.

\bibitem[\protect\citename{Bauer}2017]{bauer2017nominal}
B.L.M. Bauer.
\newblock 2017.
\newblock {\em Nominal Apposition in Indo-European: Its Forms and Functions,
  and its Evolution in Latin-Romance}.
\newblock Trends in Linguistics. Studies and Monographs [TiLSM]. De Gruyter.

\bibitem[\protect\citename{Denkowski and
  Lavie}2014]{denkowski-lavie-2014-meteor}
Michael Denkowski and Alon Lavie.
\newblock 2014.
\newblock Meteor universal: Language specific translation evaluation for any
  target language.
\newblock In {\em Proceedings of the Ninth Workshop on Statistical Machine
  Translation}, pages 376--380, Baltimore, Maryland, USA, June. Association for
  Computational Linguistics.

\bibitem[\protect\citename{Devlin \bgroup et al.\egroup }2019]{devlin2019bert}
Jacob Devlin, Ming-Wei Chang, Kenton Lee, and Kristina Toutanova.
\newblock 2019.
\newblock Bert: Pre-training of deep bidirectional transformers for language
  understanding.
\newblock In {\em Proceedings of the 2019 Conference of the North American
  Chapter of the Association for Computational Linguistics: Human Language
  Technologies, Volume 1 (Long and Short Papers)}, pages 4171--4186.

\bibitem[\protect\citename{Gu \bgroup et al.\egroup
  }2016]{gu-etal-2016-incorporating}
Jiatao Gu, Zhengdong Lu, Hang Li, and Victor~O.K. Li.
\newblock 2016.
\newblock Incorporating copying mechanism in sequence-to-sequence learning.
\newblock In {\em Proceedings of the 54th Annual Meeting of the Association for
  Computational Linguistics (Volume 1: Long Papers)}, pages 1631--1640, Berlin,
  Germany, August. Association for Computational Linguistics.

\bibitem[\protect\citename{Joulin \bgroup et al.\egroup
  }2018]{joulin-etal-2018-loss}
Armand Joulin, Piotr Bojanowski, Tomas Mikolov, Herv{\'e} J{\'e}gou, and
  Edouard Grave.
\newblock 2018.
\newblock Loss in translation: Learning bilingual word mapping with a retrieval
  criterion.
\newblock In {\em Proceedings of the 2018 Conference on Empirical Methods in
  Natural Language Processing}, pages 2979--2984, Brussels, Belgium,
  October-November. Association for Computational Linguistics.

\bibitem[\protect\citename{Kang \bgroup et al.\egroup }2019]{kang2019pomo}
Jun~Seok Kang, Robert L~Logan IV, Zewei Chu, Yang Chen, Dheeru Dua, Kevin
  Gimpel, Sameer Singh, and Niranjan Balasubramanian.
\newblock 2019.
\newblock Pomo: Generating entity-specific post-modifiers in context.
\newblock In {\em Proceedings of NAACL-HLT}, pages 826--838.

\bibitem[\protect\citename{Kim and Mooney}2010]{kim-mooney-2010-generative}
Joohyun Kim and Raymond Mooney.
\newblock 2010.
\newblock Generative alignment and semantic parsing for learning from ambiguous
  supervision.
\newblock In {\em Coling 2010: Posters}, pages 543--551, Beijing, China,
  August. Coling 2010 Organizing Committee.

\bibitem[\protect\citename{Konstas and Lapata}2013]{Konstas2013AGM}
Ioannis Konstas and Mirella Lapata.
\newblock 2013.
\newblock A global model for concept-to-text generation.
\newblock {\em J. Artif. Intell. Res.}, 48:305--346.

\bibitem[\protect\citename{Liang \bgroup et al.\egroup
  }2009]{liang-etal-2009-learning}
Percy Liang, Michael Jordan, and Dan Klein.
\newblock 2009.
\newblock Learning semantic correspondences with less supervision.
\newblock In {\em Proceedings of the Joint Conference of the 47th Annual
  Meeting of the {ACL} and the 4th International Joint Conference on Natural
  Language Processing of the {AFNLP}}, pages 91--99, Suntec, Singapore, August.
  Association for Computational Linguistics.

\bibitem[\protect\citename{Nivre \bgroup et al.\egroup
  }2020]{nivre-etal-2020-universal}
Joakim Nivre, Marie-Catherine de~Marneffe, Filip Ginter, Jan Haji{\v{c}},
  Christopher~D. Manning, Sampo Pyysalo, Sebastian Schuster, Francis Tyers, and
  Daniel Zeman.
\newblock 2020.
\newblock {U}niversal {D}ependencies v2: An evergrowing multilingual treebank
  collection.
\newblock In {\em Proceedings of the 12th Language Resources and Evaluation
  Conference}, pages 4034--4043, Marseille, France, May. European Language
  Resources Association.

\bibitem[\protect\citename{Papineni \bgroup et al.\egroup
  }2002]{10.3115/1073083.1073135}
Kishore Papineni, Salim Roukos, Todd Ward, and Wei-Jing Zhu.
\newblock 2002.
\newblock Bleu: A method for automatic evaluation of machine translation.
\newblock In {\em Proceedings of the 40th Annual Meeting on Association for
  Computational Linguistics}, ACL ’02, page 311–318, USA. Association for
  Computational Linguistics.

\bibitem[\protect\citename{P{\k{e}}zik}2016]{pkezik2016exploring}
Piotr P{\k{e}}zik.
\newblock 2016.
\newblock Exploring phraseological equivalence with paralela.

\bibitem[\protect\citename{Qi \bgroup et al.\egroup }2020]{qi2020stanza}
Peng Qi, Yuhao Zhang, Yuhui Zhang, Jason Bolton, and Christopher~D. Manning.
\newblock 2020.
\newblock Stanza: A {Python} natural language processing toolkit for many human
  languages.
\newblock In {\em Proceedings of the 58th Annual Meeting of the Association for
  Computational Linguistics: System Demonstrations}.

\bibitem[\protect\citename{Rothe \bgroup et al.\egroup
  }2019]{DBLP:journals/corr/abs-1907-12461}
Sascha Rothe, Shashi Narayan, and Aliaksei Severyn.
\newblock 2019.
\newblock Leveraging pre-trained checkpoints for sequence generation tasks.
\newblock {\em CoRR}, abs/1907.12461.

\bibitem[\protect\citename{Tiedemann}2012]{TIEDEMANN12.463}
Jörg Tiedemann.
\newblock 2012.
\newblock Parallel data, tools and interfaces in opus.
\newblock In Nicoletta Calzolari~(Conference Chair), Khalid Choukri, Thierry
  Declerck, Mehmet~Ugur Dogan, Bente Maegaard, Joseph Mariani, Jan Odijk, and
  Stelios Piperidis, editors, {\em Proceedings of the Eight International
  Conference on Language Resources and Evaluation (LREC'12)}, Istanbul, Turkey,
  may. European Language Resources Association (ELRA).

\bibitem[\protect\citename{Traffis}2019]{traffis_2019}
Catherine Traffis.
\newblock 2019.
\newblock Appositives-what they are and how to use them, May.

\bibitem[\protect\citename{Vrande\v{c}i\'{c} and
  Kr\"{o}tzsch}2014]{10.1145/2629489}
Denny Vrande\v{c}i\'{c} and Markus Kr\"{o}tzsch.
\newblock 2014.
\newblock Wikidata: A free collaborative knowledgebase.
\newblock {\em Commun. ACM}, 57(10):78–85, September.

\bibitem[\protect\citename{Yamada \bgroup et al.\egroup
  }2017]{yamada-etal-2017-learning}
Ikuya Yamada, Hiroyuki Shindo, Hideaki Takeda, and Yoshiyasu Takefuji.
\newblock 2017.
\newblock Learning distributed representations of texts and entities from
  knowledge base.
\newblock {\em Transactions of the Association for Computational Linguistics},
  5:397--411.

\end{thebibliography}

\clearpage
\appendix

\section{Appendices}

\subsection{Manual validation} 
\label{manual}
We hired annotators fluent in the languages of the data to manually validate it. They had to mark instances where an error had occurred in the appositive detection, the detected appositive was not factual, or the entity had been incorrectly linked to WikiData (all instances of \textit{noise} in the data, resulting from errors in appositive detection and entity linking). Our ultimate goal was to build test sets of 1,000 instances per language per entity type, equally balanced between positive and negative instances, i.e. we needed 500 valid data instances per language per entity type. Based on a pilot study, we determined that noise levels for candidate appositives for \textsc{person} entities were approximately 33\% and for \textsc{organization} entities, 50\% (averaged across languages). We therefore gave annotators 750 and 1,000 candidates to annotate for the \textsc{person} and \textsc{organization} types, respectively. For most language-entity type combinations, the manual annotation successfully yielded close to 500 valid instances. That was not the case for Polish \textsc{organization} appositives, where only 80 valid candidates were retrieved, so we excluded this language-entity type combination from our work. It remains an open question whether \textsc{organization} appositives in Polish are rare or our automatic detection method failed at catching them. 

\subsection{Composition of cross-lingual data}
\label{composition}
Table~\ref{tab:facttypes} present findings on the composition of the Spanish, German and Polish positions of the data, respectively, as observed through cross-referencing with WikiData. The low number of facts for Polish is the result of one fact type dominating a large amount of the data (\textit{position held}).

\begin{table}[]
    \begin{subtable}{0.45\textwidth}
    \centering
    \small
    \begin{tabular}{clll}
                         &Fact type&News (\%)&Wiki(\%)\\
                         \hline                         
                         \multirow{7}{*}{\rotatebox{90}{\textsc{per}}}&  position held & 20.9 & 9.4 \\ 
                          & occupation&15.9&10.6\\
                          & citizenship&10.1&4.3\\
                          & member of party&7.6&1.9\\
                          & award received&5.2&3.9\\
                          & nominated for&3.6&0.4\\
                          & educated at&3.1&3.1\\
                         \hline
                         \multirow{7}{*}{\rotatebox{90}{\textsc{org}}} & instance of&23.1&10.9\\
                         & official website & 6.3&6.2\\
                         & country&5.9&3.3\\
                         & member of&4.2&2.4\\
                         & subsidiary&3.5&2.1\\
                         & capital of&3.2&0.1\\
                         & has quality&3.0&0.0\\
    \end{tabular}
    \caption{English }
    \label{tab:wikidata_en}
    \end{subtable}
    \begin{subtable}{0.45\textwidth}
    \small
    \centering
    \begin{tabular}{clll}
                         &Fact type&News (\%)&Wiki(\%)\\
                         \hline                       \multirow{7}{*}{\rotatebox{90}{\textsc{per}}}&  position held&27.1&4.9\\
                          & occupation&11.2&10.3\\ 
                          & citizenship&9.8&4.8\\
                          & participant of&8.7&1.1\\
                          & member of party&7.3&1.3\\
                          & award received&4.1&3.2\\
                          & employer&3.2&0.5\\
                         \hline \multirow{6}{*}{\rotatebox{90}{\textsc{org}}} & instance of&28.9&10.5\\
                         & country&7.8&3.5\\
                         & has quality&5.1&0.1\\
                         & capital of&4.4&0.5\\
                         & member of&4.3&2.9\\
                         & is located in&3.4&2.2\\
    \end{tabular}
    \caption{Spanish }
    \label{Spanish}
    \end{subtable}
    \begin{subtable}{0.45\textwidth}
    \small
    \centering
    \begin{tabular}{clll}
                         &Fact type&News (\%)&Wiki(\%)\\
                         \hline                       \multirow{8}{*}{\rotatebox{90}{\textsc{per}}}&  position held&35.2&6.7\\
                          & employer&5.9&2.9\\ 
                          & citizenship&5.9&1.3\\
                          & member of party&5.3&2.4\\
                          & award received&4.9&2.1\\
                          & occupation&4.8&3.6\\
                          & participant of&4.6&1.5\\
                          & member of & 3.6&1.0\\
                         \hline \multirow{6}{*}{\rotatebox{90}{\textsc{org}}} & official website&15.6&12.4\\
                         & instance of&15.1&4.5\\
                         & owner of&7.5&2.4\\
                         & member of&5.8&0.9\\
                         & has quality&4.6&0.0\\
                         & Commons category&3.2&4.5\\
    \end{tabular}
    \caption{German }
    \label{German}
    \end{subtable}
    \begin{subtable}{0.45\textwidth}
    \small
    \centering
    \begin{tabular}{clll}
                         &Fact type&News (\%)&Wiki(\%)\\
                         \hline                       \multirow{2}{*}{\rotatebox{90}{\textsc{per}}}&  position held&77.2&5.7\\
                          & participant of&3.2&0.7\\

    \end{tabular}
    \caption{Polish }
    \label{Polish}
    \end{subtable}
    \caption{Top fact types per language.}
    \label{tab:facttypes}
\end{table}

\subsection{Implementations and hyperparameters}
\label{implementation}
\paragraph{[Base] and [KB]}
We use the implementation of \newcite{kang2019pomo} from \url{https://github.com/rloganiv/claimrank-allennlp}. We set the model hyperparameters to the ones reported in their paper, changing only the dimension of the embeddings from 500 to 300, to make the comparison between [Base] and [KB] fair in terms of model parameterization. Training hyperparameters were tweaked to achieve stable training that fits on one 16 GB GPU. See the full list of hyperparameters in Table~\ref{tab:hps}.

\subsection{Projection of NTEE embeddings}
\label{projection}
We obtained a bilingual dictionary with CSLS retrieval over the cross-lingual FastText embeddings. CSLS retrieval is similar to nearest neighbor retrieval, but has proven more accurate: \cite{joulin-etal-2018-loss} report an accuracy of 83.7\%, 77.6\% and 73.5\% for word translation, respectively, from Spanish, German and Polish to English, as measured on a sample of 1,500 medium frequency words. Any errors in the bilingual dictionaries would inevitably lead to noise in the NTEE embedding projection.

\paragraph{Copynet}
We use the AllenNLP \cite{Gardner2017AllenNLP} implementation of Copynet with the hyperpameters shown in  Table~\ref{tab:hps}.

\begin{table}[]
    \centering
    \small
    \begin{tabular}{lll}
         &  Base/KB & Copynet \\
         \hline
         vocab size & 50k & 50k \\
         embedding dim & 300 & 300 \\
         hidden units & 250 & 250 \\
         num layers & 2 & 2 \\
         optimizer & Adam & Adam \\
         learning rate & 0.001 & 0.0001 \\
         batch size & 16 & 6 \\
         dropout & 0.3 & 0.3 \\

    \end{tabular}
    \caption{Hyperparameter configurations for Base/KB models and Copynet models.}
    \label{tab:hps}
\end{table}

\subsection{Transformer experiments}
\label{transformer}
Transformer-based architectures are state-of-the-art for many NLP tasks, so it is only fair that we experiment with such an architecture as well. As BERT models \cite{devlin2019bert} have been made available for all four languages we work with, we chose to train BERT-to-BERT encoder-decoder models for appositive generation. \newcite{DBLP:journals/corr/abs-1907-12461} found that architecture to give strong performance on tasks like sentence fusion and rephrasing. We used their training schedule but unfortunately, found that all models learned to predict the \texttt{<empty>} token exclusively. As it is not the goal of our work to explore the capabilities of the BERT-to-BERT architecture in particular, we did not use further resources to adjust the training schedule. Yet, we do believe this to be an optimization problem, and we would not discourage future research from attempting to solve the task of appositive generation with a transformer-based approach.

\subsection{Results}
\begin{figure*}
    \centering

    \begin{subfigure}{0.39\linewidth}
    \resizebox{\linewidth}{!}{
    \includegraphics{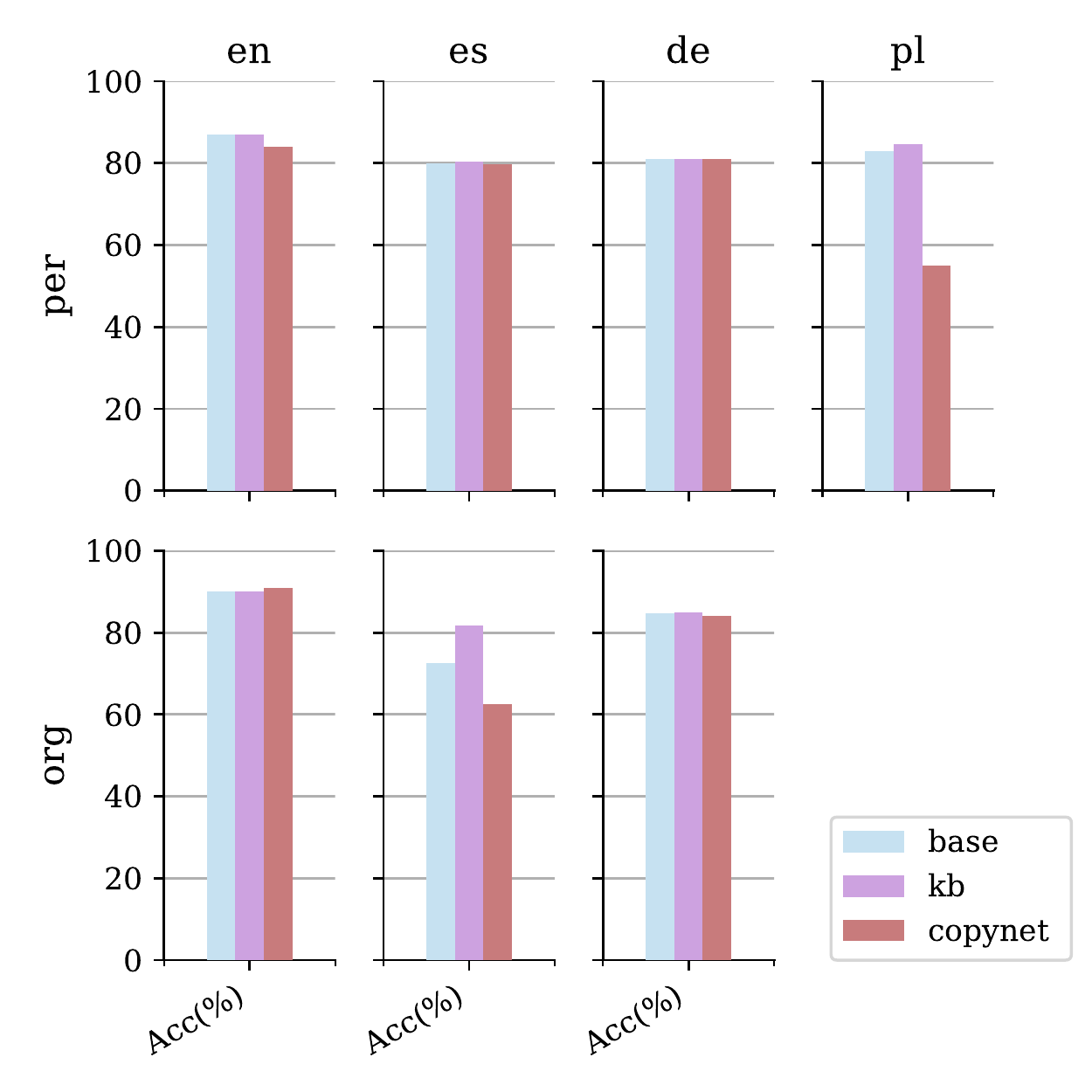}}
    \caption{}
    \end{subfigure}
    \begin{subfigure}{0.59\linewidth}
    \resizebox{\textwidth}{!}{
    \includegraphics{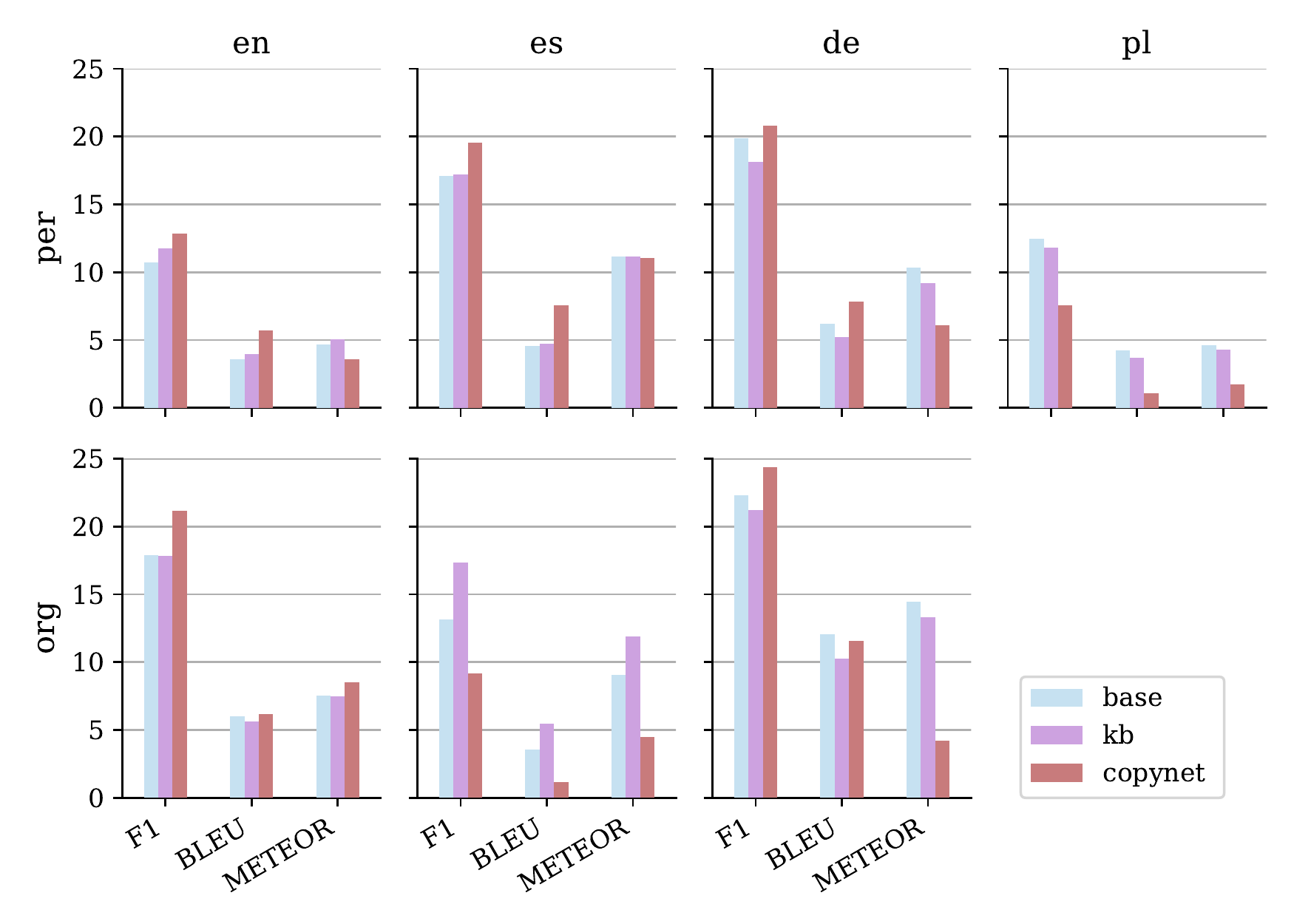}}
    \caption{}
    \end{subfigure}
    \caption{(a) Evaluation of (a) the models' ability to correctly decide when an appositive is due, (b) generated predictions for positive test instances. Measured on the News test set. }
    \label{fig:silver}
\end{figure*}

The results as measured on the Wikipedia test set are shown in Figure~\ref{fig:silver}. Compared to results on the News test set (see Figure~\ref{fig:gold}, the numbers seen here are higher, which is to be expected considering that this test set is in-domain and any noise found in it (due to it being silver standard) likely resembles the noise found in the training data. It is worth noting though, that certain patterns repeat between the two test sets, as for example the fact that \texttt{copynet}, as measured on F1 score and BLEU, outperforms the other models on the majority of language-named entity type combinations, but not on Polish \textsc{person} appositives and Spanish \textsc{organization} appositives. This suggests that, while evaluation on the silver-standard Wikipedia test set cannot be consider fully stable and representative, it can be taken as a proxy in model comparison for developmental purposes.

The numbers behind the results from Figures~\ref{fig:gold} and~\ref{fig:silver} are shown in Tables~\ref{tab:gold_results} and ~\ref{tab:silver_results}, respectively.

\begin{table*}[]
    \centering
    \begin{tabular}{cllllllll}
        && Acc & F1 & BLEU & METEOR\\
        \hline
        \multirow{3}{*}{\rotatebox{90}{\textsc{en-per}}}&always yes &0.5&0.0&0.0&0.0\\
        &base &71.91&0.72&1.03&2.96\\
        &kb &71.73&0.73&1.03&2.9\\
        &copynet &70.19&0.71&2.45&2.24\\
        \hline
        \multirow{3}{*}{\rotatebox{90}{\textsc{en-org}}}&always yes &0.5&0.0&0.0&0.0\\
        &base &68.09&0.74&0.23&1.61\\
        &kb &66.58&0.74&0.21&1.58\\
        &copynet &65.98&0.73&0.52&1.37\\
        \hline
        \multirow{3}{*}{\rotatebox{90}{\textsc{es-per}}}&always yes &0.58&0.0&0.0&0.0\\
        &base &62.84&0.67&1.24&5.44\\
        &kb &61.18&0.68&0.79&4.79\\
        &copynet &64.08&0.67&2.2&4.36\\
        \hline
        \multirow{3}{*}{\rotatebox{90}{\textsc{es-org}}}&always yes &0.4&0.0&0.0&0.0\\
        &base &47.39&0.68&1.32&5.64\\
        &kb &60.91&0.71&2.02&6.83\\
        &copynet &33.57&0.62&0.33&2.77\\
        \hline
        \multirow{3}{*}{\rotatebox{90}{\textsc{de-per}}}&always yes &0.54&0.0&0.0&0.0\\
        &base &62.94&0.67&0.8&3.19\\
        &kb &63.47&0.68&0.75&2.84\\
        &copynet &67.63&0.69&1.75&1.76\\
        \hline
        \multirow{3}{*}{\rotatebox{90}{\textsc{de-org}}}&always yes &0.46&0.0&0.0&0.0\\
        &base &62.77&0.72&0.54&3.21\\
        &kb &65.13&0.73&0.19&3.47\\
        &copynet &62.22&0.72&0.43&1.08\\
        \hline
        \multirow{3}{*}{\rotatebox{90}{\textsc{pl-per}}}&always yes &0.54&0.0&0.0&0.0\\
        &base &72.83&0.78&0.16&0.09\\
        &kb &73.8&0.77&0.06&0.13\\
        &copynet &2.75&0.58&0.0&0.12\\
    \end{tabular}
    \caption{Evaluation of the models' ability to correctly decide when an appositive is due and of generated predictions for positive test instances. Measured on the News test set. }
    \label{tab:gold_results}
\end{table*}

\begin{table*}[]
    \centering
    \begin{tabular}{cllllllll}
    \hline
    \multirow{3}{*}{\rotatebox{90}{\textsc{en-per}}}&base &85.42&0.87&3.59&4.7\\
    &kb &85.32&0.87&3.94&5.07\\
    &copynet &81.95&0.84&5.7&3.57\\
    \hline
    \multirow{3}{*}{\rotatebox{90}{\textsc{en-org}}}&base &89.47&0.9&5.99&7.54\\
    &kb &89.43&0.9&5.65&7.49\\
    &copynet &89.17&0.91&6.17&8.53\\
    \hline
    \multirow{3}{*}{\rotatebox{90}{\textsc{es-per}}}&base &78.49&0.8&4.57&11.17\\
    &kb &78.73&0.8&4.71&11.19\\
    &copynet &79.1&0.8&7.54&11.04\\
    \hline
    \multirow{3}{*}{\rotatebox{90}{\textsc{es-org}}}&base &61.63&0.73&3.58&9.07\\
    &kb &80.69&0.82&5.45&11.92\\
    &copynet &46.91&0.62&1.14&4.46\\
    \hline
    \multirow{3}{*}{\rotatebox{90}{\textsc{de-per}}}&base &79.28&0.81&6.19&10.35\\
    &kb &80.09&0.81&5.24&9.19\\
    &copynet &80.12&0.81&7.85&6.11\\
    \hline
    \multirow{3}{*}{\rotatebox{90}{\textsc{de-org}}}&base &84.09&0.85&12.04&14.45\\
    &kb &84.58&0.85&10.28&13.31\\
    &copynet &82.07&0.84&11.57&4.19\\
    \hline
    \multirow{3}{*}{\rotatebox{90}{\textsc{pl-per}}}&base &82.67&0.83&4.23&4.61\\
    &kb &84.46&0.85&3.67&4.29\\
    &copynet &24.49&0.55&1.1&1.71\\
    \end{tabular}
    \caption{Evaluation of the models' ability to correctly decide when an appositive is due and of generated predictions for positive test instances. Measured on the Wiki test set. }
    \label{tab:silver_results}
\end{table*}

\subsection{Ranking paradigm study}
\label{taste_test}
\begin{figure}
    \centering
    \resizebox{0.55\linewidth}{!}{
    \includegraphics{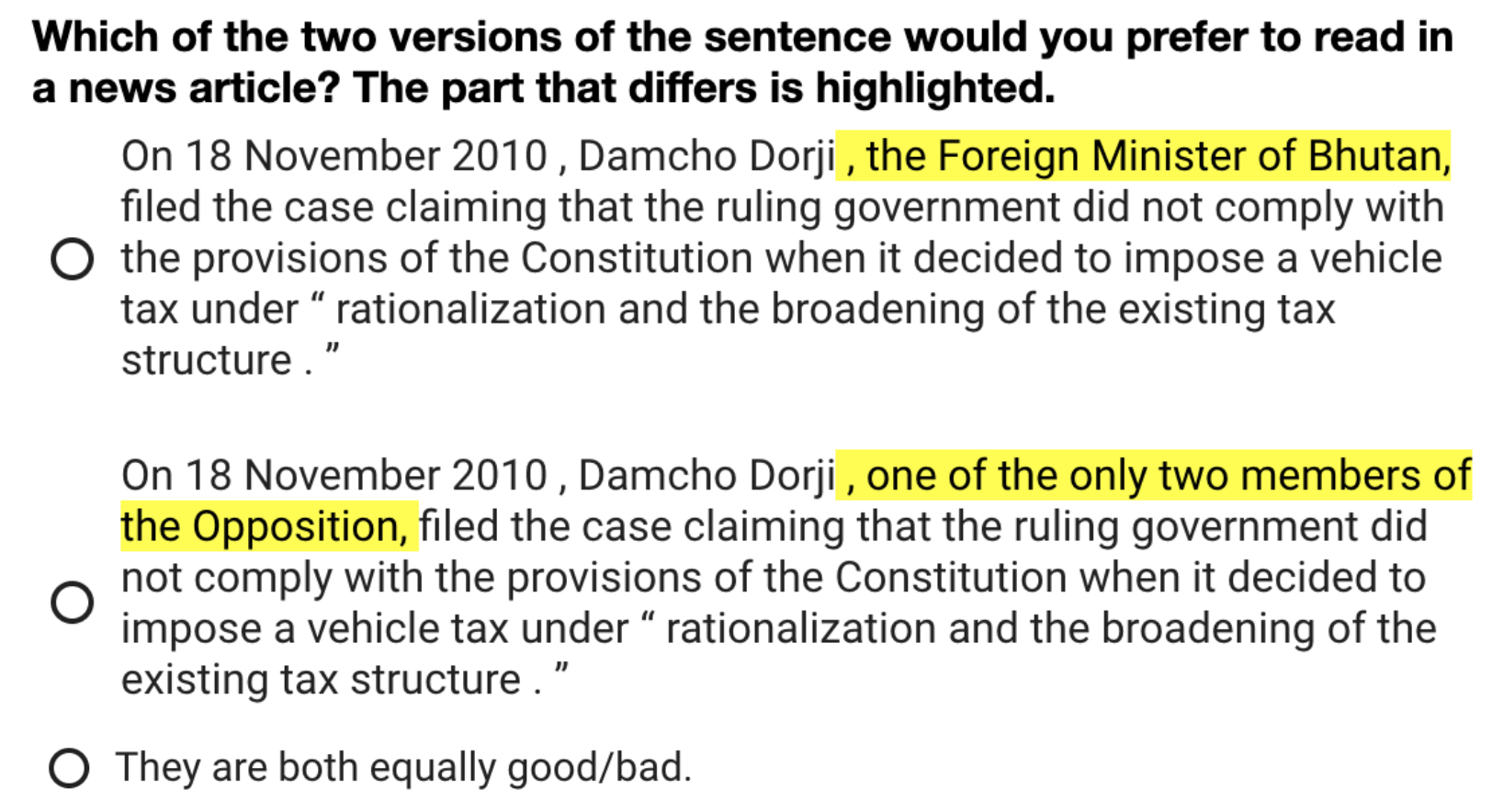}}
    \caption{Prompt for manual evaluation.}
    \label{fig:tastetest}
\end{figure}

Figure~\ref{fig:tastetest} shows an example prompt from the blind taste test. Instances where either the true appositive was empty or the predicted one was empty were included in the the study, but instances where both were empty were excluded, as the comparison would not have been meaningful in this case. The average time for completing a HIT was 53 seconds.

\end{document}